\documentclass{article}

\usepackage[preprint]{neurips_2024}

\usepackage[utf8]{inputenc} 
\usepackage[T1]{fontenc}    
\usepackage{hyperref}       
\usepackage{url}            
\usepackage{booktabs}       
\usepackage{amsfonts}       
\usepackage{nicefrac}       
\usepackage{microtype}      
\usepackage{xcolor}         
\usepackage{graphicx}
\usepackage{amsmath}
\usepackage{subcaption}

\title{Towards noise contrastive estimation with soft targets for conditional models} 

\author{%
    Johannes Hugger\\
    European Bioinformatics Institute \\
    European Molecular Biology Laboratory\\
  \texttt{jhugger@ebi.ac.uk} \\
  \And
    Virginie Uhlmann\\
    European Bioinformatics Institute \\
    European Molecular Biology Laboratory\\
    \texttt{uhlmann@ebi.ac.uk} \\
}

\begin{document}

\maketitle

\begin{abstract}
Soft targets combined with the cross-entropy loss have shown to improve generalization performance of deep neural networks on supervised classification tasks.
The standard cross-entropy loss however assumes data to be categorically distributed, which may often not be the case in practice. In contrast, InfoNCE does not rely on such an explicit assumption but instead implicitly estimates the true conditional through negative sampling.
Unfortunately, it cannot be combined with soft targets  in its standard formulation, hindering its use in combination with sophisticated training strategies.
In this paper, we address this limitation by proposing a loss function that is compatible with probabilistic targets.
Our new soft target InfoNCE loss is conceptually simple, efficient to compute, and can be motivated through the framework of noise contrastive estimation.
Using a toy example, we demonstrate shortcomings of the categorical distribution assumption of cross-entropy, and discuss implications of sampling from soft distributions.
We observe that soft target InfoNCE performs on par with strong soft target cross-entropy baselines and outperforms hard target NLL and InfoNCE losses on popular benchmarks, including ImageNet.
Finally, we provide a simple implementation of our loss, geared towards supervised classification and fully compatible with deep classification models trained with cross-entropy.
\end{abstract}

\section{Introduction}
\label{submission}
The cross-entropy loss, or log loss, and its soft target variants are amongst the most popular objective functions for conditional density estimation in supervised classification problems~\citep{he2016deep, dosovitskiy2020image}.
In its base variant, it assumes a degenerate one-hot distribution $p(l|x)=\delta(l,k)$ as the underlying conditional density over labels $l$ given data $x$, which might not suffice to capture the nature of complex or ambiguously annotated datasets.
This simplifying assumption can have a negative effect on parameter estimation as well as model calibration properties~\citep{guo2017calibration, muller2019does}, and adversely affects the generalization capability of the model.
To mitigate this, training with probabilistic targets, also called soft targets, has shown to be beneficial~\citep{szegedy2016rethinking, muller2019does}.
For instance, label smoothing injects noise into labels, leading to a higher entropy in the target conditional distribution~\citep{szegedy2016rethinking}.
This was shown to be effective in a wide range of problems and, when combined with other soft target techniques~\citep{zhang2017mixup, han2022g}, leads to state-of-the-art performance~\citep{dosovitskiy2020image, he2022masked}.

Several avenues have been proposed to improve upon the log loss~\citep{gutmann2010noise, lin2017focal}, including the InfoNCE objective~\citep{jozefowicz2016exploring, ma2018noise}.
InfoNCE has become the standard loss function for self-supervised contrastive learning due to its information theoretic properties, simplicity and excellent empirically-observed performance~\citep{chen2020simple, he2020momentum, tian2020contrastive, oord2018representation}.
It has shown to produce versatile, high-quality representations that optimize a lower bound on their mutual information (MI) content~\citep{tian2020contrastive, oord2018representation}.
In natural language processing, InfoNCE was shown to be an efficient replacement for the log loss, with better generalization performance and on-par parameter estimation properties~\citep{jozefowicz2016exploring, ma2018noise}.
InfoNCE avoids the summation over all labels that is usually required to compute the normalization constant and instead relies on negative sampling, which is be advantageous when label spaces are large.
Negative sampling has a strong connection to noise contrastive estimation as it ultimately estimates the true conditional $p(k|x)$ implicitly instead of using an explicit model~\citep{gutmann2010noise, ma2018noise}.

Despite its favourable theoretical properties as well as its empirical success in natural language processing and, more broadly, self-supervised learning, InfoNCE has not seen a wide adoption in supervised classification tasks involving other types of data such as images or graphs.
One reason for this might be the difficulty of 
identifying scenarios in which InfoNCE would be a favourable loss function.
Another reason might be that, due to its noise contrastive nature, combining it with powerful methods leveraging soft targets~\citep{szegedy2016rethinking, zhang2017mixup, han2022g, yun2019cutmix} is not straightforward. This contrasts with the log loss, where soft target probabilities can be explicitly used to model the target conditional density.
It is however conceivable that integrating such targets and their corresponding distributions might boost generalization performance.

In this work, we first show that InfoNCE is a relevant alternative to the log loss, particularly in the case of label uncertainty.
We demonstrate through a toy example that InfoNCE indeed yields better parameter estimates in this scenario.
This motivates us to suggest approaches that combine soft targets with noise contrastive estimation.
Our main contribution is an InfoNCE generalization that directly integrates soft target probabilities into the loss by creating weighted linear combinations of label embeddings.
Our loss function, which we dub \emph{soft target InfoNCE}, can be motivated through noise contrastive estimation by leveraging the continuous categorical distribution in combination with a Bayesian argument.
We discuss the resulting attraction-repulsion dynamics as well as how this prior affects gradients.
As second approach we suggest to sample targets from a distribution with higher entropy, such as the label smoothing distribution $p^\epsilon(y|x)=(1-\epsilon) p(y|x) + \epsilon \xi(y)$ in which $\epsilon \in [0, 1]$ determines the weighting between the underlying conditional and the noise distribution.
Furthermore, we investigate how this affects the MI and entropy in the learned representation by deriving a variational lower bound.

Finally, we perform experiments on multiple classification benchmarks, including ImageNet, to compare and examine soft target InfoNCE's properties.
Our results suggest that our loss is competitive with soft target cross-entropy and outperforms hard target NLL and InfoNCE.
In an ablation study we give insights into the confidence calibration of the trained models and investigate how the number of negative samples and batch size, as well as the amount of label smoothing noise, affects model performance.
Our results confirm the intuition that InfoNCE and soft target InfoNCE are important alternatives to the log loss and its soft variant.
Our implementation is available at \url{https://github.com/uhlmanngroup/soft-target-InfoNCE.git}.

\section{Background and related work}

\subsection{Log loss, soft targets and the continuous categorical distribution}\label{sec:intrologloss}

In classification problems, the log loss, also known as negative log likelihood (NLL) or cross entropy loss, is an objective to evaluate the quality of probabilistic models~\citep{hastie2009elements}. It is the standard loss function used to train deep supervised classification models~\citep{krizhevsky2017imagenet, he2016deep, dosovitskiy2020image}.
Intuitively, given a data sample, the log loss quantifies the discrepancy between the predicted class probability of the true label and the probability given through a target distribution.
The most commonly used target is the degenerate conditional distribution $p(l|x)=\delta(l,k)$, induced by the one-hot encoding of labels into probabilities.
The estimated probabilities $q$ are usually computed as a softmax over class scores $q(k|x, \theta):= \exp(z \cdot w_k) / \sum_{l=1}^K \exp(z \cdot w_l)$, where $k$ is the class label, $z$ denotes the penultimate layer activations (also called embedding) of sample $x$, and $w_k$ corresponds to the weights of the classifier layer responsible for computing the respective score.
The log loss function can then be defined for sample $x$ with label $k$ as
\begin{align} \label{eq:nll}
    l^{\mathrm{NLL}}(x,k)&:= - \log q(k|x, \theta).
\end{align}
A number of works have addressed different drawbacks of the log loss, including its robustness to noisy labels~\citep{zhang2018generalized, sukhbaatar2014training} and poor margins~\citep{elsayed2018large, cao2019learning}.
In addition to these shortcomings, neural networks trained with~\eqref{eq:nll} tend to be poorly calibrated, which leads to overconfident predictions and to an increased misclassification rate~\citep{guo2017calibration}.
Besides scaling logits with a temperature parameter, label smoothing has shown to mitigate this problem~\citep{muller2019does}.
Label smoothing has been introduced as a model regularization technique in which the original "hard" one-hot encoding of a label is replaced by a "soft" target, computed as the weighted average of the original hard target and a noise distribution over labels, for instance the uniform distribution $\xi \equiv 1 / K$~\citep{szegedy2016rethinking}.
Formally, the label smoothing distribution can be defined as
\begin{align} \label{eq:softtargetce_optimum}
    p^\epsilon(l|x) = (1-\epsilon) \delta (l,k) + \epsilon \xi(l),
\end{align}
where $\xi$ denotes a noise distribution, $\epsilon \in [0, 1]$, and where the delta distribution $\delta(l,k) = p(l|x)$ is the original target conditional distribution.
During training, \eqref{eq:nll} is replaced by the soft target cross-entropy given by
\begin{align} \label{eq:SoftTargetCE}
    l^\mathrm{SoftTargetNLL}(x,k) =
    -\sum_{l=1}^K p^\epsilon(l|x) \log q(l|x, \theta).
\end{align}
By distributing small probability mass over labels that do not correspond to the true class, models are incentivised to reduce their prediction confidence, which has shown to improve performance~\citep{dosovitskiy2020image}.
Besides classification, soft targets are also used in knowledge distillation, in which a teacher network provides a student network its estimated probabilities as training targets~\citep{hinton2015distilling}.

The idea of distributing probability masses over two or more labels has also been used in combination with mixing input samples in a data augmentation technique called MixUp~\citep{zhang2017mixup}.
In MixUp data points, including their labels, are blend together into a new sample using a mixing coefficient, resulting in a new sample and a corresponding soft target.
In this paper, we refer to targets that distribute probability mass among multiple classes as soft targets regardless of the underlying sample. 
This includes data augmentation techniques such as MixUp, as well as regularization methods like label smoothing.

Recently,~\cite{gordon2020uses} proposed a label smoothing loss based on the continuous categorical distribution~\citep{gordon2020continuous} as an alternative to~\eqref{eq:SoftTargetCE} that models non-categorical data more faithfully.
The continuous categorical distribution generalizes its discrete analog and is defined on the closed simplex as
\begin{align}
    \alpha_1, \dots, \alpha_K \sim \mathcal{CC}(\lambda) \ \
    \Longleftrightarrow \ \ p(\alpha_1, \dots, \alpha_K; \lambda) \propto \prod_{i=1}^K \lambda_i^{\alpha_i},
\end{align}
with $\sum_{i=1}^K \alpha_i = 1$.

\subsection{InfoNCE for supervised classification}

InfoNCE has been demonstrated to be a versatile loss function with applications across many machine learning tasks and application domains~\citep{tian2020contrastive, he2020momentum, poole2019variational, damrich2022contrastive, zimmermann2021contrastive}.
In this work, we focus predominantly on its density estimation properties.
In order to estimate densities, InfoNCE turns the problem into a supervised classification task in which the position of a single sample coming from the data distribution $p$ must be identified in a tuple $T$ of otherwise noise samples coming from a distribution $\eta$~\citep{jozefowicz2016exploring}.
The theoretical footing for conditional density estimation and classification has been introduced in the context of language modelling~\citep{ma2018noise}.
Concretely,~\cite{ma2018noise} shows that InfoNCE's optimum $\theta^*$ are the parameters of the true conditional, \emph{i.e.} $q(k|x, \theta^*) = p(k|x)$.
In other words, supervised InfoNCE fits $\theta$ by minimizing the negative log likelihood of $(x, k)$ being at a position $S$ within $T$, given by $-\log P(S|T)$.
Without loss of generality, for $S=0$, the InfoNCE loss can then be defined as
\begin{align} \label{eq:infonce}
    l^{\mathrm{InfoNCE}}(T)=
    -\log \frac{\exp(s(z, y_{0k_0}) / \tau)}{\sum_{i=0}^N \exp(s(z, y_{ik_i}) / \tau)},
\end{align}
with $(x, y_{0k_0}) \sim p, y_{1k_1}, \dots, y_{Nk_N} \sim \eta$, and where $\tau$ denotes the temperature and $s(z,y;\tau, \eta) = z^T y / \tau - \log \eta(y)$ is the scoring function. 
Further, $y_{ik_i}$ is the learnable label embedding of class $k_i$, and $i$ denotes the position of $(x, k_i)$ within $T$.
The label embeddings are a parametrized lookup table and are the analogs to the classifier scoring weights $w_k$ of a neural network trained with the log loss.

Several variants of InfoNCE have been introduced, mostly in the context of self-supervised learning~\citep{chuang2020debiased, kalantidis2020hard, chen2020simple, li2023rethinking}.
The supervised variant proposed by~\cite{khosla2020supervised} is used for pre-training and leverages labels to create positive pairs. Further variants developed by~\cite{chuang2020debiased} and~\cite{li2023rethinking} correct for false negatives coming from the noise distribution by re-weighting the sum over negative samples, while~\cite{kalantidis2020hard} proposes to sample and create artificial hard negatives for training.

InfoNCE is also a biased estimator of MI~\citep{oord2018representation}. Maximizing the MI between learned representations that are based on related inputs (InfoMax principle,~\cite{linsker1988self}) has yielded empirical success in representation learning~\citep{hjelm2018learning, henaff2020data, bachman2019learning, chen2020simple, he2020momentum, velivckovic2018deep}.
As MI is notoriously difficult to estimate, tractable variational lower bounds are usually optimized in practice~\citep{poole2019variational}.
InfoNCE provides such a lower bound: more specifically, optimizing InfoNCE is equivalent to maximizing a lower bound of MI~\citep{oord2018representation, tian2020contrastive}.

\section{Results}

\subsection{InfoNCE improves parameter estimation over negative log-likelihood in conditional models}
Precise parameter estimation is essential for conditional models to obtain good generalization properties. 
Here we compare the estimation accuracy of InfoNCE against NLL in classification settings with different degrees of label uncertainty.

In its standard formulation, NLL makes the assumption that the target conditional distribution is degenerate, \emph{i.e.} $p(l|x)=\delta(l, k)$ for a data sample $x$ with class label $k$.
However, in practice the true conditional distribution is often more complex, for instance due to uncertainty or ambiguity in the label assignment.
As a consequence, an inherent amount of error is introduced into the estimated parameters.
Several works have addressed this issue, for instance by introducing more entropy into the reference distribution~\cite{szegedy2016rethinking} or through alternative loss functions~\cite{gutmann2010noise}, including InfoNCE.
InfoNCE does not require an explicit model as target distribution.
Instead, parameters are estimated in an implicit manner by contrasting labels coming from the true distribution with labels coming from a noise distribution.

To study the benefits of this implicit approach, we consider cases in which the underlying conditional has different amounts of uncertainty (Figure Supp. \ref{fig:tsne}).
Specifically, we sample data points $x$ from a mixture of Gaussians with $K$ modes.
Each mode $\theta_k \in \mathbb{R}^d$ corresponds to a class in the sense that labels for the data points $x$ are sampled from the conditional model given by
\begin{align}
    p(k|x, \theta) = \frac{\exp(x^T \theta_k)}{\sum_{l=1}^K \exp(x^T \theta_l)}.
\end{align}
Concretely, we first sample $\mathcal{X}=\{x_1, \dots, x_M\}$ from the mixture distribution and then subsample $\mathcal{X}$ into a dataset $X$ of size $N > M$ with duplicates.
Class labels are assigned by sampling from the conditional model $k \sim p(\cdot | x, \theta)$.
We then estimate the parameters $\theta_k$ using NLL and InfoNCE.
We increase the uncertainty by increasing the degree of alignment between modes $\theta_k$, as detailed in Supplementary Section \ref{supp:density_estimation}.

\begin{figure}[tb]
  \centering
  \includegraphics[height=6.5cm]{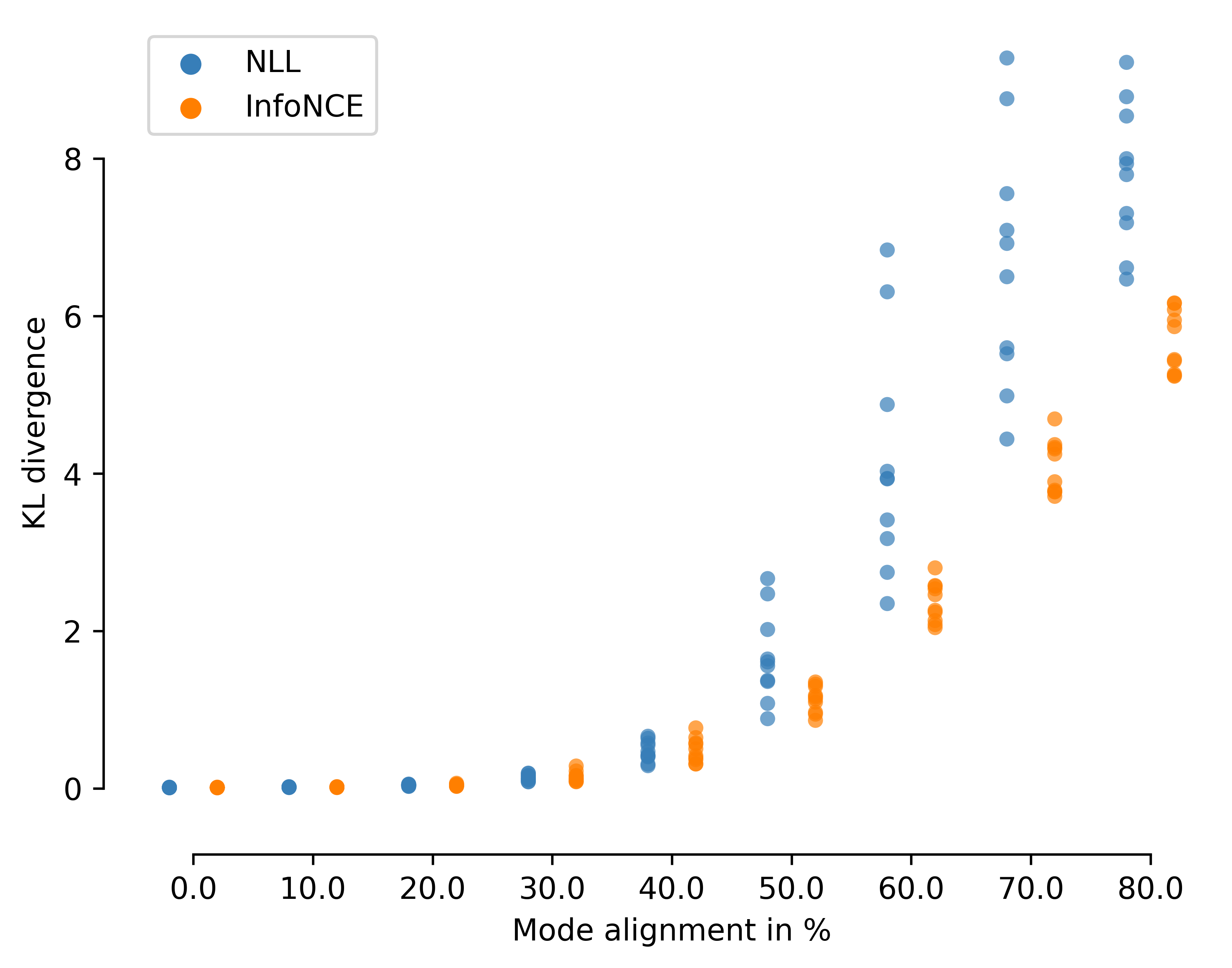}
  \caption{Parameter estimation quality of InfoNCE vs. negative log-likelihood in conditional density models with different degrees of mode alignment.
The estimation error is reported as the KL divergence. The alignment degree corresponds to the angle between modes. $0\%$ alignment corresponds to orthogonal modes and $80\%$ corresponds to an angle of $18^\circ$.
  }
  \label{fig:density_estimation}
\end{figure}

The results, visualized in Figure~\ref{fig:density_estimation}, indicate that increased levels of label uncertainty lead to worse parameter estimates using NLL when compared to InfoNCE.
As anticipated, parameter estimation accuracy deteriorates as the mode alignment degree increases for both loss functions.
However, the error of InfoNCE's estimates relative to those of NLL increases considerably more slowly.
Our experiment demonstrates the advantage of implicit sampling from a dataset generated by the true data distribution as opposed to using an explicit model distribution as target.

In order to connect this result with deep neural networks, one can interpret the data $X$ as embeddings (\emph{i.e.} penultimate layer activations) and the $\theta_k$ as the classification weights producing the logits of the last layer.
For orthogonal modes ($0 \%$ alignment), $X$ is linearly separable and both objectives give comparable estimates.
However, for non-orthogonal modes (alignment $>0\%$), InfoNCE might lead to better generalization.
This makes our observations particularly relevant in fine-tuning settings where only the last layer is re-trained.

\subsection{InfoNCE with soft targets} \label{sec:SoftTargetInfoNCE}

Motivated by InfoNCE's favourable parameter estimation properties and the empirical success of training with soft targets, we sought to combine both.
In this section, we propose a conceptually simple loss function that integrates soft targets directly into InfoNCE.
We begin by motivating it through the derivation of a theoretically sound objective function, which we then simplify to overcome a computational bottleneck in the sampling process.

\paragraph{Derivation.}
Consider the continuous categorical distributions $\mathcal{CC}(p_{Y|X}(\cdot | x))$, $\mathcal{CC}(\eta)$ induced by the true conditional and noise distribution, respectively.\footnote{To simplify notation we denote $\mathcal{CC}([p_{Y|X}(1 | x), \dots, p_{Y|X}(K | x)])$ as $\mathcal{CC}(p_{Y|X}(\cdot | x))$ and $\mathcal{CC}([\eta_1, \dots, \eta_K])$ as $\mathcal{CC}(\eta)$}
Let $T=(\alpha_1, \dots, \alpha_{N+1})$ be a tuple of length $N+1$, with $\alpha_i \geq 0$, $\sum_{l=1}^K \alpha_{il} = 1$, containing a single sample $\alpha \sim \mathcal{CC}(p_{Y|X}(\cdot | x))$ and noise samples $\tilde{\alpha} \sim \mathcal{CC}(\eta)$ otherwise.
We are interested in identifying the position $S$ of $\alpha$ given the tuple $T$, \emph{i.e.} $P(S|T)$.
A priori, we have $P(S=i) = 1 / (N+1)$ for all allowed positions $i = 1, \dots, N+1$.
Further, we know that $P(\alpha_j | S=k) = p_\mathcal{CC}(\alpha_j; \eta)$ for all $j \neq k$, and $P(\alpha_k | S=k) = p_\mathcal{CC}(\alpha_k; p_{Y|X}(\cdot | x))$. Thus,
\begin{align} \label{eq:stnce_likelihood}
    P(T|S=k) &= p_\mathcal{CC}(\alpha_k; p_{Y|X}(\cdot | x)) \prod_{\substack{i=1 \\ i \neq k}}^{N+1} p_\mathcal{CC}(\alpha_i; \eta) 
             = \frac{p_\mathcal{CC}(\alpha_k; p_{Y|X}(\cdot | x))}{p_\mathcal{CC}(\alpha_k; \eta)} \prod_{i=1}^{N+1} p_\mathcal{CC}(\alpha_i; \eta).
\end{align}
From this, we can infer the marginal
\begin{align} \label{eq:stnce_marginal}
    P(T) = \frac{1}{N+1} 
     \prod_{i=1}^{N+1} p_\mathcal{CC}(\alpha_i; \eta)
     \sum_{k=1}^{N+1}
    \frac{p_\mathcal{CC}(\alpha_k; p_{Y|X}(\cdot | x))}{p_\mathcal{CC}(\alpha_k; \eta)}.
\end{align}
By applying Bayes' theorem in combination with~\eqref{eq:stnce_likelihood} and~\eqref{eq:stnce_marginal}, we can compute the posterior
\begin{align} \label{eq:stnce_posterior}
    P(S=k | T) = \frac{
    \frac{p_\mathcal{CC}(\alpha_k; p_{Y|X}(\cdot | x))}{p_\mathcal{CC}(\alpha_k; \eta)}
    }{
    \sum_{l=1}^{N+1}\frac{p_\mathcal{CC}(\alpha_l; p_{Y|X}(\cdot | x))}{p_\mathcal{CC}(\alpha_l; \eta)}
    }
    =  \frac{
    \prod_{i=1}^K \left( \frac{p_{Y|X}(i|x)}{\eta_i} \right)^{\alpha_{ki}}
    }{
    \sum_{l=1}^{N+1} \prod_{j=1}^K \left( \frac{p(_{Y|X}(j|x)}{\eta_j}
    \right)^{\alpha_{lj}}
    }.
\end{align}
We estimate the density ratio $p_{Y|X}(i|x) / \eta_i$ using the parametric model $\exp(s(z, y_k; \tau, \eta))$ where $s(z, y_k; \tau, \eta) = z^Ty_k / \tau - \log \eta_{y_k}$ and $z=f(x; \theta)$, which can for instance be a neural network.
By inserting this back into~\eqref{eq:stnce_posterior}, we can state the maximum likelihood estimation problem
\begin{align} \label{eq:stnce_mle}
    \max_{\theta}
    \frac{
     \exp \left( \sum_{i=1}^K \alpha_{ki} s(z, y_i; \tau, \eta)\right) 
    }{
    \sum_{l=1}^{N+1} \exp \left( \sum_{j=1}^K \alpha_{lj} s(z, y_j; \tau, \eta) \right)
    }.
\end{align}
Taking the negative logarithm, we can rewrite~\eqref{eq:stnce_mle} as the 
loss function
\begin{align} \label{eq:theo-soft-target-infonce}
    \underset{
    \substack{\alpha_k \sim \mathcal{CC}(p_{Y|X}) \\ \alpha_1, \dots, \alpha_{k-1}, \alpha_{k+1}, \dots, \alpha_{N+1} \sim \mathcal{CC}(\eta)}
    }{\mathbb{E}}
    \left[
    - \log
      \frac{
    \exp \left(\sum_{i=1}^K \alpha_{ki} s(z, y_i; \tau, \eta) \right)
    }{
    \sum_{l=1}^{N+1} \exp \left( \sum_{j=1}^K \alpha_{lj} s(z, y_j; \tau, \eta) \right)
    }
    \right].
\end{align}
Since $\exp (\sum_{i=1}^K \alpha_{ki} s(z, y_i; \tau, \eta))$ converges to the density ratio $p(\alpha_l; p_{Y|X}) / p(\alpha_l; \eta)$, we can simply use $\exp(z^Ty_i / \tau)$ during inference to estimate a proportional score of the probability $p_{Y|X}(i|x)$.

\paragraph{Computational considerations.}
Unfortunately, the loss function~\eqref{eq:theo-soft-target-infonce} cannot directly be used in practice since sampling from the continuous categorical distribution is inefficient, especially with many classes (\emph{e.g.}, $K=1000$).
Thus, instead of sampling $\alpha_l$, we opt to directly use class probabilities, which demonstrated good empirical performance.
More specifically, we set $\alpha_k = [p_{Y|X}(1|x), \dots, p_{Y|X}(K|x)]$ and leverage soft targets from other batch samples as negatives, \emph{i.e.}, we set $\alpha_l=[\eta_{l1}, \dots, \eta_{lK}]$, $l\neq k$.
Incorporating these changes, we can rewrite~\eqref{eq:theo-soft-target-infonce} as our soft target InfoNCE loss function
\begin{align} \label{eq:soft-target-infonce}
    \underset{
    \substack{x \sim p_X \\ \eta_1, \dots, \eta_N \sim p_Y}
    }{\mathbb{E}}
    \left[
    - \log
      \frac{
    \exp \left(\sum_{i=1}^K p_{Y|X}(i|x) s(z, y_i; \tau, \eta) \right)
    }{
    \exp \left(\sum_{i=1}^K p_{Y|X}(i|x) s(z, y_i; \tau, \eta) \right) + \sum_{l=1}^{N} \exp \left( \sum_{j=1}^K \eta_{lj} s(z, y_j; \tau, \eta) \right)
    }
    \right].
\end{align}

\paragraph{Discussion.}
To provide an intuition for~\eqref{eq:soft-target-infonce}, we restate it as a negative temperature scaled energy cross-entropy loss
\begin{align} \label{eq:energyXent}
    - \sum_{i=1}^K p_{Y|X}(i|x) 
    \log
      \frac{
    \exp \left( s(z, y_i; \tau, \eta) \right)
    }{
     \exp \left(\sum_{i=1}^K p_{Y|X}(i|x) s(z, y_i; \tau, \eta) \right) + \sum_{l=1}^{N} \exp \left( \sum_{j=1}^K \eta_{lj} s(z, y_j; \tau, \eta) \right)
    },
\end{align}
where the fraction inside of the logarithm can be interpreted as an energy function\footnote{Note that $q(k|z)=\exp(s(z, y_i)) / \left(\exp \left(\sum_{i=1}^K p_{Y|X}(i|x) s(z, y_i) \right) + \sum_{l=1}^{N} \exp \left( \sum_{j=1}^K \eta_{lj} s(z, y_j) \right)\right)$ is not necessarily a probability since $\sum_{k=1}^K q(k|x)$ does not sum to 1 in general.}.
Notably,~\eqref{eq:energyXent} is structurally similar to the soft target cross-entropy~\eqref{eq:SoftTargetCE}.
For each class, we now have an attractive pull acting on its embedding $y_i$ and $z$ weighted by soft targets. 
For instance, in case of label smoothing with a uniform noise distribution $\xi \equiv 1 / K$, we have $p(i|x) = 1 - (K-1/K) \epsilon$ for the true class $i$ and $p(j|x) = \epsilon / K$ otherwise, which results in a dampened attraction between the data embedding and the original hard target.

We keep the inference model $\exp(z^Ty_i / \tau)$ of the original InfoNCE, but note that the correction term $-\log \eta$, used during training, might introduce additional bias.
Furthermore, we note that the gradient dymanics induced by the new energy landscape are different from standard InfoNCE, and result in a changed attraction-repulsion spectrum between embeddings.
A more detailed discussion of this change as well as a derivation of the gradient is provided in Supplementary Section~\ref{supp:snce_discussion}.

\subsection{InfoNCE with soft distributions} \label{sec:sdInfoNCE}

An alternative to integrating soft weights directly into the loss function is to sample from the corresponding distribution instead.
In this section, we examine an InfoNCE loss in which labels are drawn from the soft distribution
\begin{align} \label{eq:softdistribution}
    p^\epsilon(k|x) = (1-\epsilon) p(k|x) + \epsilon \xi(k),
\end{align}
instead of the data generated by the true conditional.
A derivation of this loss is provided in Supplementary Section~\ref{supp:sdinfonce} and follows directly from the standard derivation of InfoNCE~\citep{damrich2022contrastive}.

\paragraph{Effect of soft distributions on the MI lower bound.}

Here, we sketch the derivation and discuss the implications of sampling from the soft distribution~\eqref{eq:softdistribution} on InfoNCE's MI lower bound with respect to the underlying data distribution $p$.
We provide a complete proof in Supplementary Section~\ref{supp:sdinfonce}.

As we saw in the previous section, the optimal critic is proportional to the density ratio between the soft data and the noise distribution
\begin{align}
    \exp(s(z, y) / \tau) & \propto \frac{p_{Y|Z}^\epsilon(y|z)}{p_Y(y)},
\end{align}
with $\eta \equiv \xi \equiv p_Y$.
Using the argument provided in the proof of the MI lower bound~\citep{tian2020contrastive, oord2018representation}, we directly obtain
\begin{align} \label{eq:MI_prebound}
    l^\mathrm{SDInfoNCE}(T) \geq \log N -
    \underset{(y, z) \sim p^\epsilon}{\mathbb{E}} \left[ \log \frac{p_{Y|Z}^\epsilon(y|z)}{p_Y(y)} \right].
\end{align}
The expected value can be written as
\begin{align}
    &\underset{(y, z) \sim p^\epsilon}{\mathbb{E}} \left[ \log \frac{p_{Y|Z}^\epsilon(y|z)}{p_Y(y)} \right]
    =I(Z;Y) + H(Y|Z) - H_\epsilon(Y|Z),
\end{align}
where $H(Y|Z)$ denotes the conditional entropy with respect to $p_{Y,Z}$ and
\begin{align}
    H_\epsilon(Y|Z)=
    \underset{(y, z) \sim p_{Y,Z}^\epsilon}{\mathbb{E}} \left[ - \log p_{Y|Z}^\epsilon(y|z) \right].
\end{align}
Inserting this back into~\eqref{eq:MI_prebound} yields following inequality
\begin{align}\label{eq:softMIbound}
    I(Z;Y) + H(Y|Z) - H_\epsilon(Y|Z) \geq \log N - l^{\mathrm{SDInfoNCE}}(T).
\end{align}
For $\epsilon = 0$, we have $H_0(Y|Z)=H(Y|Z)$ and the above bound returns to its original form. In contrast, for $\epsilon = 1$, $H_1(Y|Z)=H(Y) = I(Y;Z) + H(Y|Z)$,~\eqref{eq:softMIbound} becomes trivial due to the high entropy in $p^{\epsilon=1}$.

The left hand side of~\eqref{eq:softMIbound} is larger or equal to zero (see Supplementary Section~\ref{supp:sdinfonce}).
Thus, by minimizing InfoNCE when sampling from a soft distribution, the baseline MI as well as the entropy increase.
Intuitively, $\epsilon$ regulates the amount of entropy and MI distilled into the learned representations.
High values can increase prediction uncertainty, which may help with ambiguous instances and model calibration, while low values boost model confidence, and increase MI between input and class embeddings.

\section{Experiments}

We show that soft target InfoNCE achieves competitive results compared to soft target cross-entropy and surpasses NLL, InfoNCE and soft distribution InfoNCE in quantitative evaluation on classification benchmarks, including ImageNet.
We provide an ablation study on the sensitivity to varying numbers of noise samples, varying amounts of label smoothing noise and examine calibration properties, in order to obtain further insights into the models trained with our loss.

\paragraph{Loss implementation.}
We provide a simple implementation of (soft target) InfoNCE that directly works on class scores (logits) and, thus, is compatible with standard classification models
(Figure \ref{fig:SoftTargetInfoNCE_code}).
InfoNCE loss implementations usually take as inputs embedding vectors~\citep{he2020momentum, chen2020simple, khosla2020supervised}, requiring additional code adaptations of the neural network in order to work in the supervised classification case.
More specfically, in order to compute class scores (and soft target embeddings), classification head weights need to be accessed.
While this is inconvenient, it also creates discrepancies between training and inference code.
Instead, we leverage soft targets, or one-hot encodings in case of hard targets, to directly work on the logits.

\begin{figure}[htbp]
    \centering
    \begin{verbatim}
    class SoftTargetInfoNCE(nn.Module):
    def __init__(self, noise_probs, T=1.0):
        # noise_probs: (K,) noise probabilities over classes
        # T: temperature
        super().__init__()
        self.K = noise_probs.shape[0]
        self.log_noise = torch.log(noise_probs).unsqueeze(0)
        self.T = T
    
    def forward(self, logits, targets):
        # logits: (N, K) class scores (not true logits!)
        # targets: (N, K) (soft) target weights for each class
        N = targets.shape[0]
        targets = targets.repeat(N,1).unsqueeze(1).reshape(N, N, self.K)
        logits = (logits / self.T) - self.log_noise
        logits = (logits.unsqueeze(1) * targets).sum(-1)
        logits -= (torch.max(logits, dim=1, keepdim=True)[0]).detach()
        labels = torch.arange(N, dtype=torch.long).cuda()
        return nn.CrossEntropyLoss()(logits, labels)
    \end{verbatim}
    \caption{Soft target InfoNCE PyTorch code. For clarity, this assumes training on one GPU. In the distributed case, the above has to be slightly adapted to gather (soft) targets from other GPUs to be used as additional negatives. For more details see the code repository.}
    \label{fig:SoftTargetInfoNCE_code}
\end{figure}

\subsection{Classification accuracy}

We perform image classification experiments on ImageNet \citep{krizhevsky2012imagenet}, Tiny ImageNet \citep{krizhevsky2009learning} and CIFAR-100 \citep{krizhevsky2009learning}, and node classification experiments on the CellTypeGraph benchmark \citep{cerrone2022celltypegraph}.
For image classification we use a vision transformer base (ViT-B/16,~\cite{dosovitskiy2020image}) with a 3-layer MLP classification head, pre-trained as a masked autoencoder~\citep{he2022masked} with image patches of size $16 \times 16$.
For node classification, we use a DeeperGCN~\citep{li2019deepgcns, li2020deepergcn} and report performance on this task as the average over a $5$-fold cross-validation experiment~\citep{cerrone2022celltypegraph}.

For the experiments to be comparable, we use the same training recipe and only adapt the (base) learning rate for optimal performance of the respective loss function.
Specifically, we use the same neural network architecture, fix the temperature to $\tau=1.0$, and use the dot-product as scoring function $s(z,y)=z^T y$.
For image classification, we follow common practice for ViT end-to-end fine-tuning, using the recipe provided in~\cite{he2022masked}. We train ViTs for $100$ epochs with a batch size of $1024$ using label smoothing $\epsilon=0.1$, MixUp $\alpha_M=0.8$ and CutMix of $\alpha_C=1.0$.
For node classification, we train a 32-layer DeeperGCN with a batch size of $8$ and a learning rate of $0.001$. The hard target cross-entropy and InfoNCE baselines are trained  for $500$ epochs due to overfitting, while the soft target baselines are trained for $1000$ epochs with label smoothing of $\epsilon=0.1$.
Further technical details are provided in the Supplementary Section~\ref{supp:experiments}.

\paragraph{Comparison to soft target cross-entropy.}
Table~\ref{tab:soft_target_results} shows a comparison of soft target InfoNCE to its cross-entropy loss analog across the four benchmark datasets.
While soft target cross-entropy gives slightly better results on ImageNet ($\Delta 0.29\%$) and CellTypeGraph ($\Delta0.55 \%$), we obtain on-par performance on Tiny ImageNet ($\Delta0.19 \%$) and CIFAR-100 ($\Delta0.06 \%$).
More broadly, we observe similar convergence behaviour and performance for both loss functions as a response to learning rates within the same range.
For instance, for image classification, we found that base learning rates between $0.0002$ and $0.0006$ lead to optimal results in both cases.
Notably, for image classification, we chose a $3$ layer MLP as classification head over a linear layer since we observed gains in top-$1$ accuracy for both loss functions.
However, gains for soft target InfoNCE were consistently larger ($\Delta1.72 \%$ Tiny ImageNet, $\Delta0.7 \%$ CIFAR-100) compared to soft target cross-entropy (Tiny ImageNet $\Delta0.6 \%$, CIFAR-100 $\Delta0.36 \%$) not only in top-$1$ performance but also across runs.
Since performance seems to saturate around the same accuracy levels, we conclude that soft target InfoNCE is competitive with soft target cross-entropy.

\paragraph{Comparison to InfoNCE with soft distribution sampling.}
 Table~\ref{tab:soft_target_results} compares soft target InfoNCE against the more simplistic approach of sampling a hard target from the distribution provided by the soft targets.
 Integrating soft targets directly into InfoNCE consistently results in better top-$1$ accuracy across all benchmarks.
 However, sampling from a soft distribution results in the worst performing models on CIFAR-100 and CellTypeGraph and is only on-par with the hard target baselines.

\paragraph{Comparison to hard target baselines.}
We finally compare soft target InfoNCE to the standard cross-entropy loss (NLL) and InfoNCE, which both leverage hard targets. Regularization techniques such as label smoothing and MixUp, thus, cannot be used in combination with these baselines.
While NLL is on-par with soft target InfoNCE on CIFAR-100 ($\Delta0.04 \%$) and CellTypeGraph ($\Delta0.2\%$), it is outperformed on ImageNet ($\Delta1.19 \%$) and TinyImageNet ($\Delta1.23\%$). This also holds for the InfoNCE baseline, which has the same performance pattern as the NLL baseline. 

\begin{table}[tb]
  \caption{Top-1 classification accuracy for various datasets.
  We compare soft target InfoNCE to different baseline loss functions.
  }
  \label{tab:soft_target_results}
  \centering
  \begin{tabular}{@{}lcccc@{}}
    \toprule
     & ImageNet \ \ & Tiny ImageNet \ \ & CIFAR-100 \ \ & CellTypeGraph\\
    \midrule
    NLL & 82.35 & 82.63 & \textbf{90.84} & 86.92 \\
    Soft target cross-entropy & \textbf{83.85} & 83.67 & 90.74 & \textbf{87.67} \\
    InfoNCE & 82.52 & 82.72 & 90.75 & 86.80 \\
    \midrule
    Soft distribution InfoNCE & 82.96 & 82.77 & 89.82 & 86.62 \\
    Soft target InfoNCE  & 83.54 & \textbf{83.86} & 90.80  & 87.12 \\
  \bottomrule
  \end{tabular}
\end{table}

\subsection{Ablation experiments}
In our ablation experiments, we explore different aspects of the soft target InfoNCE loss using the ViT-B/16 model described in the previous section.

\begin{figure}[tb]
  \centering
      \begin{subfigure}[b]{0.45\textwidth}
        \centering
        \includegraphics[height=4.35cm]{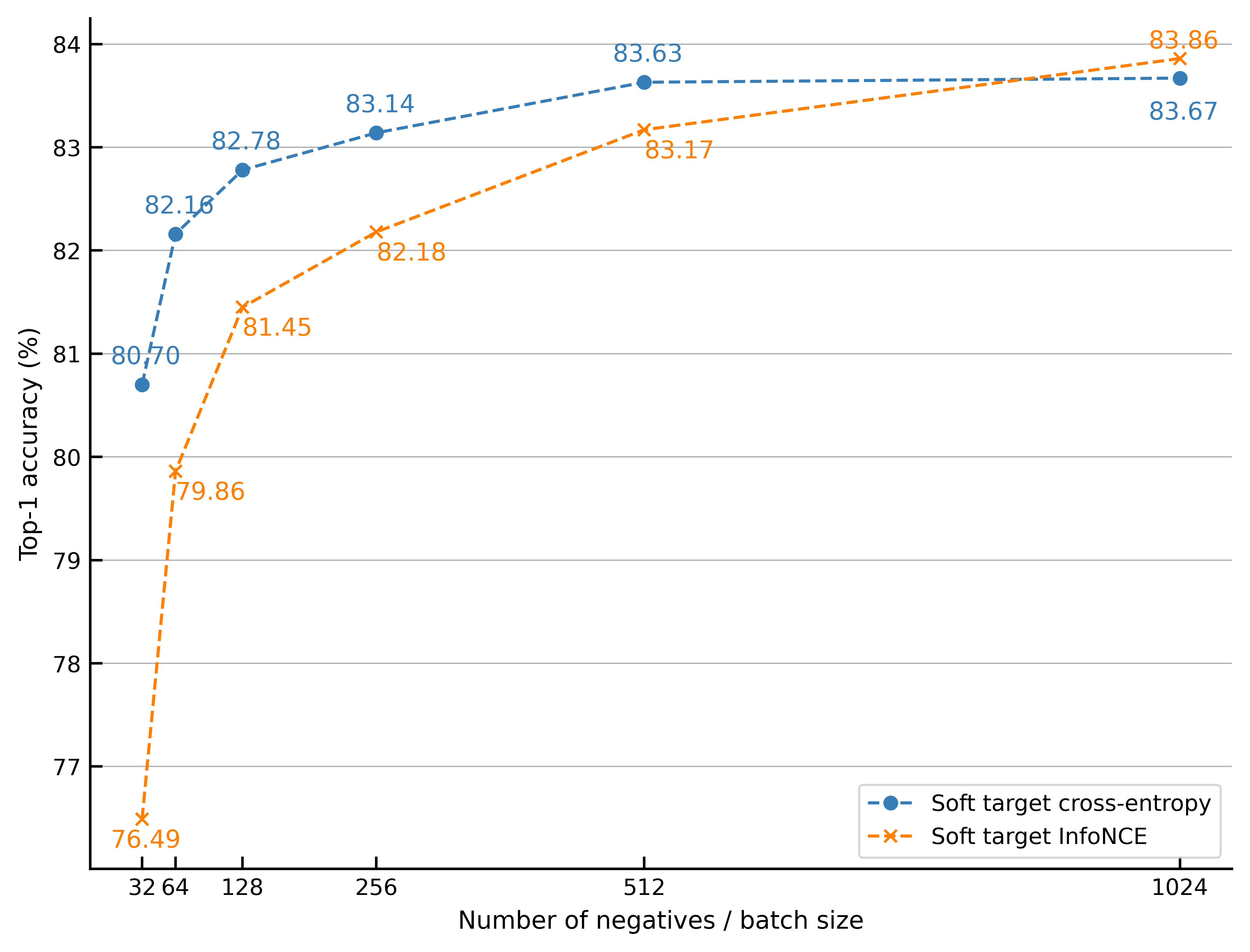}
        \caption{Top-1 accuracy as a function of noise samples/batch size for ViT-B/16 models trained with soft target losses in combination with label smoothing, MixUp and CutMix.}
    \end{subfigure}
    \hfill
    \begin{subfigure}[b]{0.45\textwidth}
        \centering
        \includegraphics[height=4.35cm]{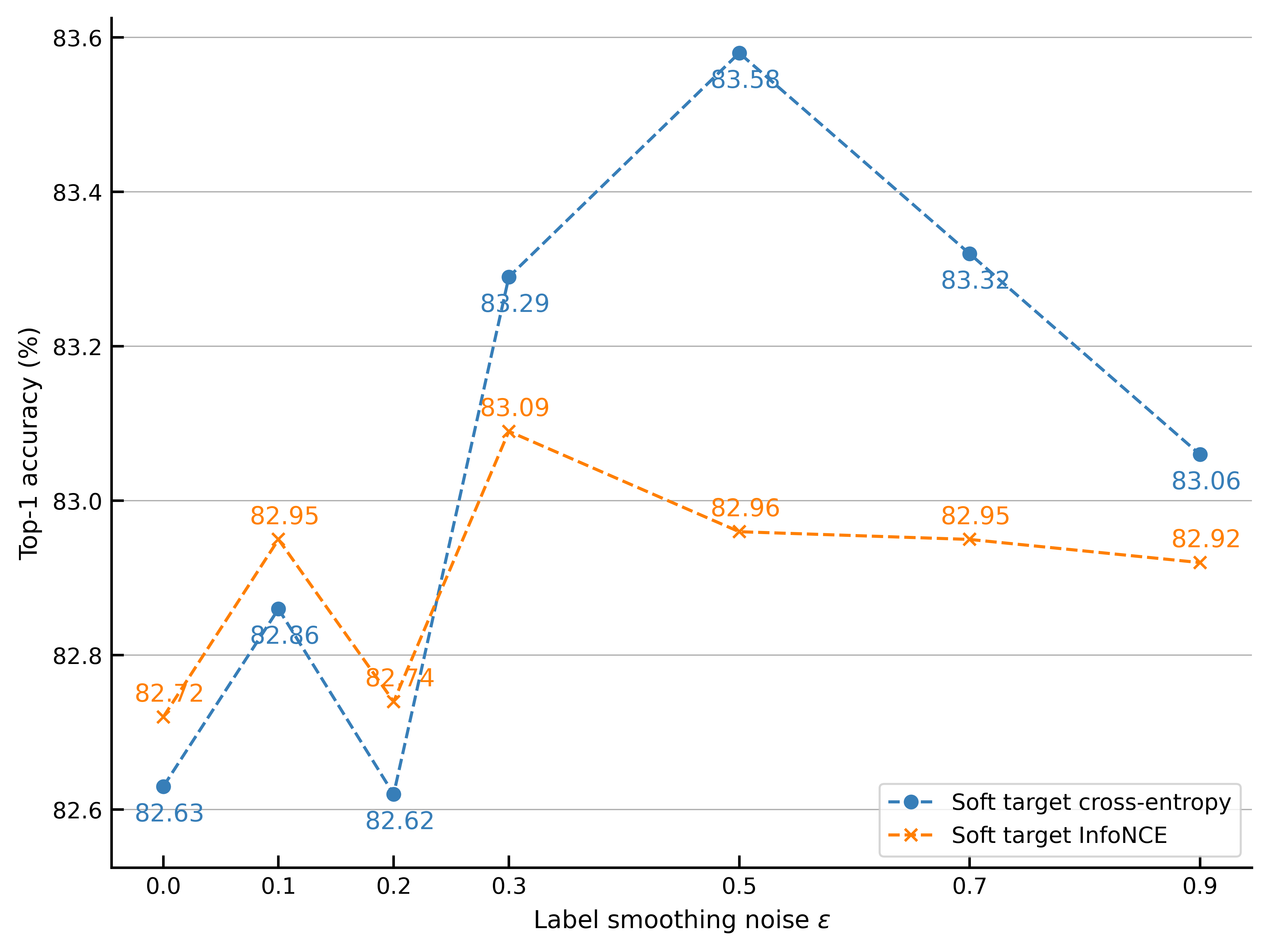}
        \caption{Top-1 accuracy as a function of label smoothing noise for ViT-B/16 models trained with soft target losses.}
    \end{subfigure}
    \caption{Noise sample (left) and label smoothing (right) ablation experiments on Tiny ImageNet.}
  \label{fig:bsz_noise_ablation}
\end{figure}

\paragraph{Number of noise samples.}
Soft target InfoNCE relies on noise samples to contrast the true soft target with.
While this can be separated from the batch size, it is more efficient to couple both and leverage other soft targets as noise samples.
We give a soft target cross-entropy baseline for comparison.
Figure~\ref{fig:bsz_noise_ablation} visualizes the results of this experiment in terms their top-1 performance on Tiny ImageNet.
As anticipated, smaller batch sizes have a stronger negative effect on performance for models trained with the noise contrastive loss compared to the baseline, due to fewer available negative samples, resulting in a performance difference of $\Delta 4.21 \%$ for the smallest batch size.
However, as the batch size increases ViTs trained with soft target InfoNCE are on-par with their cross-entropy analog, and slightly outperform them by $\Delta 0.19\%$.

\paragraph{Sensitivity to label smoothing.}
We investigate how different amounts of label smoothing noise $\epsilon$ affect models trained with soft target InfoNCE.
Figure~\ref{fig:bsz_noise_ablation} visualizes the results in terms of their top-1 accuracy on Tiny ImageNet.
Notably, accuracy of soft target InfoNCE models varies less (max deviation of $\Delta 0.37 \%$ and standard deviation of $0.12$) compared to the soft target cross-entropy baseline (max deviation of $\Delta 0.96 \%$ and standard deviation of $0.34$).

\begin{figure}[tb]
  \centering
      \begin{subfigure}[b]{0.45\textwidth}
        \centering
        \includegraphics[height=4.3cm]{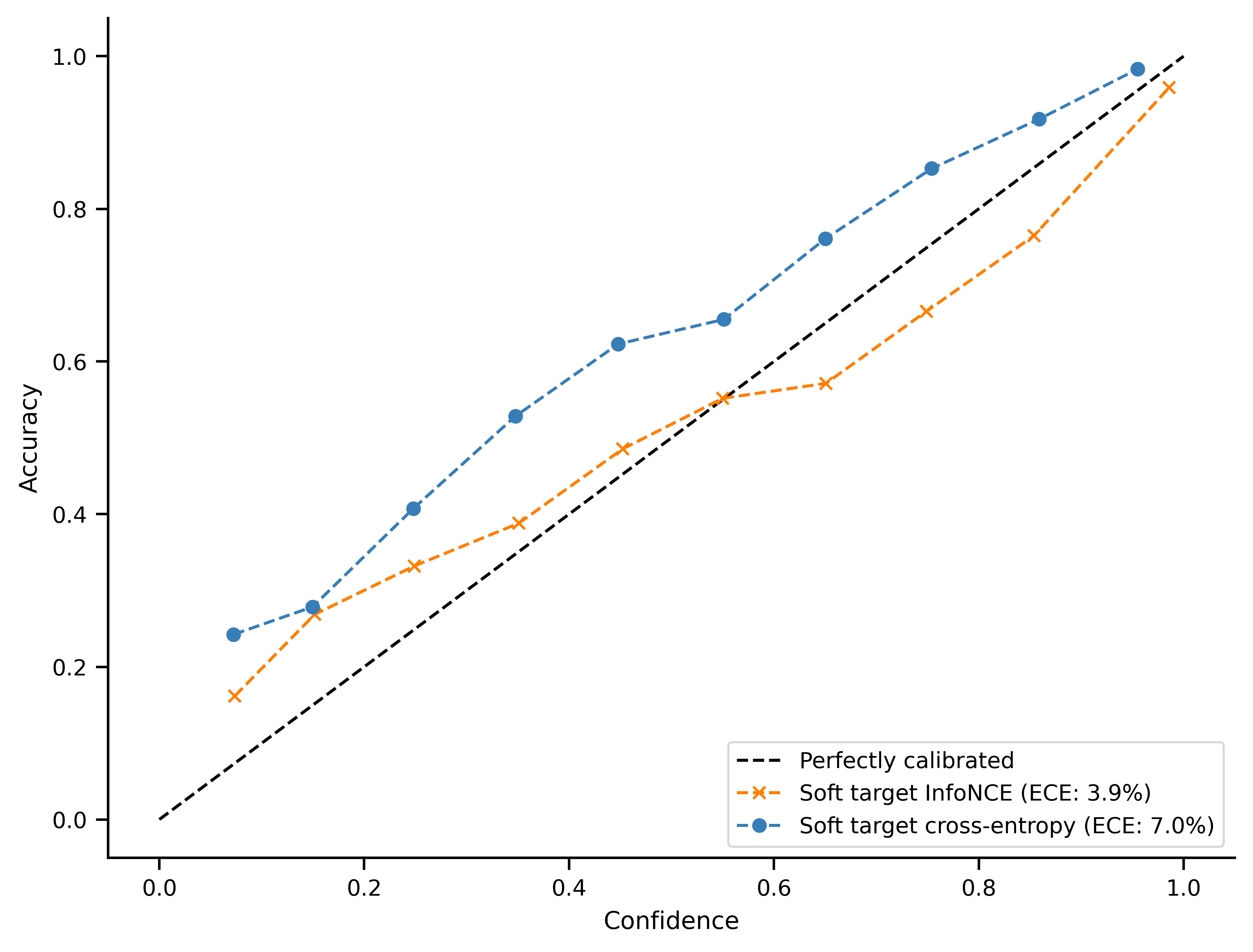}
        \caption{Soft target losses trained with label smoothing on Tiny ImageNet.}
        \label{fig:sub1}
    \end{subfigure}
    \hfill
    \begin{subfigure}[b]{0.45\textwidth}
        \centering
        \includegraphics[height=4.3cm]{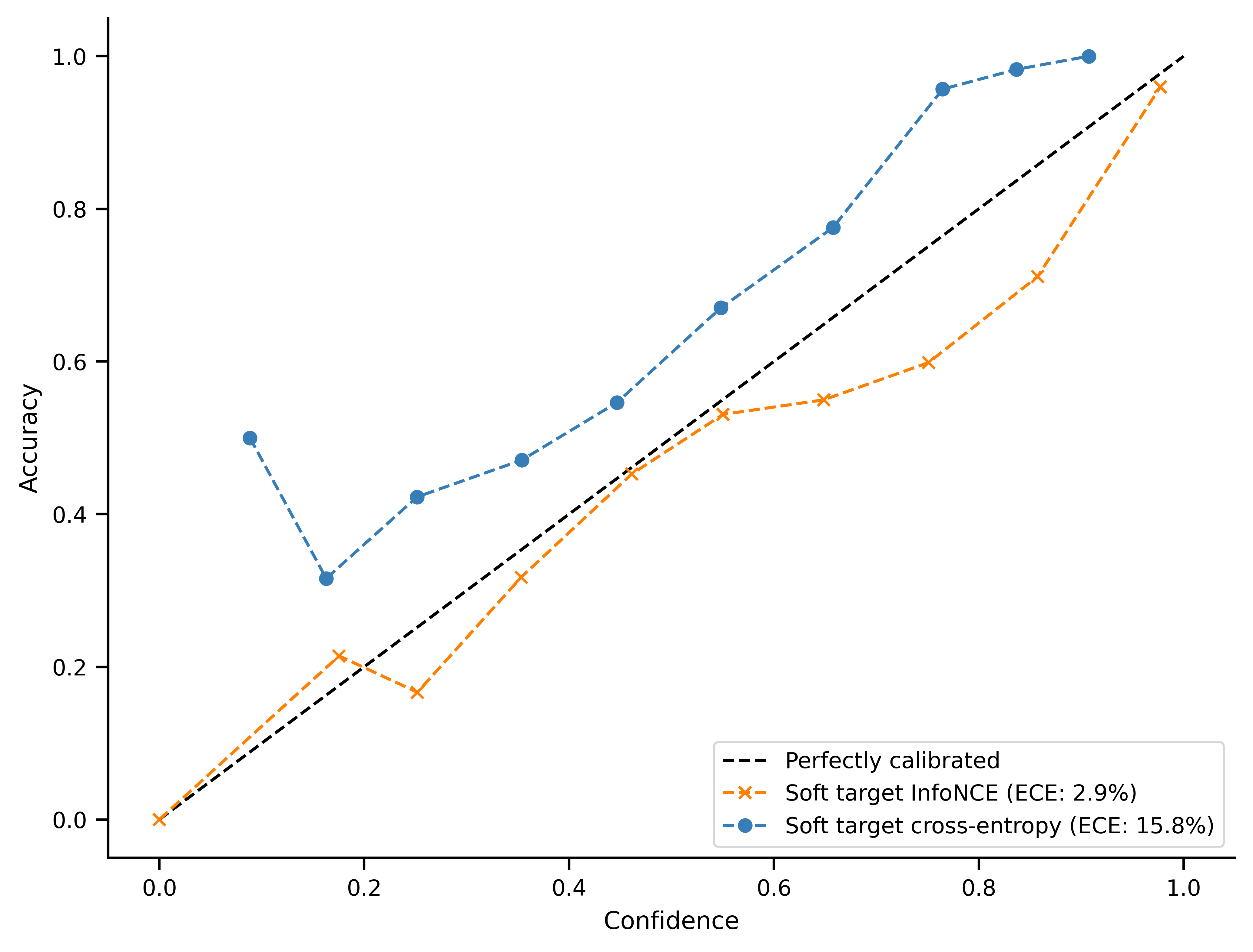}
        \caption{Best performing models on CIFAR-100 trained with soft target losses in combination with label smoothing, MixUP and CutMix.}
        \label{fig:sub2}
    \end{subfigure}
    \caption{Reliability diagrams of ViT-B/16 trained on Tiny ImageNet (left) and CIFAR-100 (right).}
  \label{fig:calibration}
\end{figure}

\paragraph{Confidence calibration.} Label smoothing in combination with cross-entropy has shown to improve model calibration~\citep{muller2019does}.
Here, we investigate confidence calibration properties of our vision transformer models trained with soft target InfoNCE and compare to its cross-entropy analog.
Since InfoNCE models converge to unnormalized density functions for which standard calibration metrics can not be computed, we treat the model outputs as logits and normalize them using a softmax function.
Figure~\ref{fig:calibration} shows reliability diagrams of ViTs trained on Tiny ImageNet with label smoothing $\epsilon=0.1$ and of the best performing ViTs on CIFAR-100 as described in the previous section.
Notably, models trained with soft target InfoNCE were better calibrated in these two cases.
Specifically, we obtained an expected calibration error (ECE) of $3.9\%$ on Tiny ImageNet and $2.9 \%$ on CIFAR-100 which was lower than the respective ECE of the soft target cross-entropy models ($7\%$ Tiny ImageNet, $15.8 \%$ CIFAR-100).
We additionally compared the best performing Tiny ImageNet models of the previous section for which we obtained a slightly worse ECE of $7.7 \%$ for the soft target InfoNCE ViT compared to an ECE of $6.3 \%$ for the soft target cross-entropy ViT (Figure \ref{supp:experiments}).

\section{Conclusions and limitations}

We presented soft target InfoNCE 
for conditional density estimation in supervised classification problems.
We motivated our loss through noise contrastive estimation, using a continuous categorical distribution in combination with a Bayesian argument.
The ability to leverage soft targets coming from sophisticated augmentation and regularization techniques allows our loss function be on-par with soft target cross-entropy and to outperform hard target NLL as well as InfoNCE losses.

As limitations we identified the drop in performance when a small batch size is used in combination with mini-batch and noise sample coupling.
However, we leave it up to future work to investigate if this can be mitigated with separately sampled soft noise targets.
Further, we observed a slight performance drop with linear classification heads when compared to soft target cross-entropy.
Nevertheless, deep classification models are typically optimized to work well with cross-entropy, and we observed competitive performance when employing a 3-layer MLP classification head.
This suggests that by exploring different neural architectures one might find more fitting alternatives.

In future work we want to explore distributions, other than the continuous categorical distribution, that can be efficiently sampled, as well as to investigate our loss in other problem settings such as knowledge distillation.

\section{Acknowledgements}

The authors thank Anna Kreshuk (EMBL) for many valuable exchanges and insightful discussions.

\bibliographystyle{plainnat}
\bibliography{bibliography}

\newpage

\appendix

\section{Supplementary material: Noise contrastive estimation with soft targets for conditional models}

\subsection{InfoNCE improves parameter estimation over negative log-likelihood in conditional models} \label{supp:density_estimation}

In order to compare parameter estimation quality of InfoNCE and NLL, we create nine datasets sampled from different Gaussian mixture models with different degrees of mode alignment.
Specifically, we sample each dataset $\mathcal{X}$ from a GMM $p(x) = \sum_{i=1}^{20} \pi_i \mathcal{N}(x|10 \cdot \theta_i, I)$, with $\theta_i \in \mathbb{R}^{20}, \|\theta_i\| = 1$ and $\pi_i \equiv 1 / 20$.
The amount of mode alignment in $\%$ in a GMM corresponds to the scaling factor applied to $\pi / 2$ determining the angles $\phi_1, \dots, \phi_{19}$ of the $20$-spherical coordinate system.
Concretely, let $s$ be a scaling factor determined by the percentage according to $s = 100 / (100 - \text{perc.})$ then $\phi$ is computed as $\phi = \pi / (2s)$.
Intuitively, we can interpret this as fixing one mode in an orthogonal system and contracting/aligning the remaining modes to the fixed one.

For each experiment we create a dataset $\mathcal{X}$ composed of $1600$ data points sampled from a GMM.
$\mathcal{X}$ is then uniformly subsampled into $X$ with $|X|=32000$, and we assign labels to each $x \in X$ based on the conditional model $p(k|x, \theta) = \exp(x^T \theta_k) / \sum_{l=1}^K \exp(x^T \theta_l)$ described in the main text.

For fitting the parameters, we train for $500$ epochs with early stopping, a batch size of $1024$ and a learning rate of $0.001$.
We evaluate parameter estimation quality for each GMM over 10 random seeds.

\begin{figure}[t]
 \centering
 \includegraphics[width=\columnwidth]{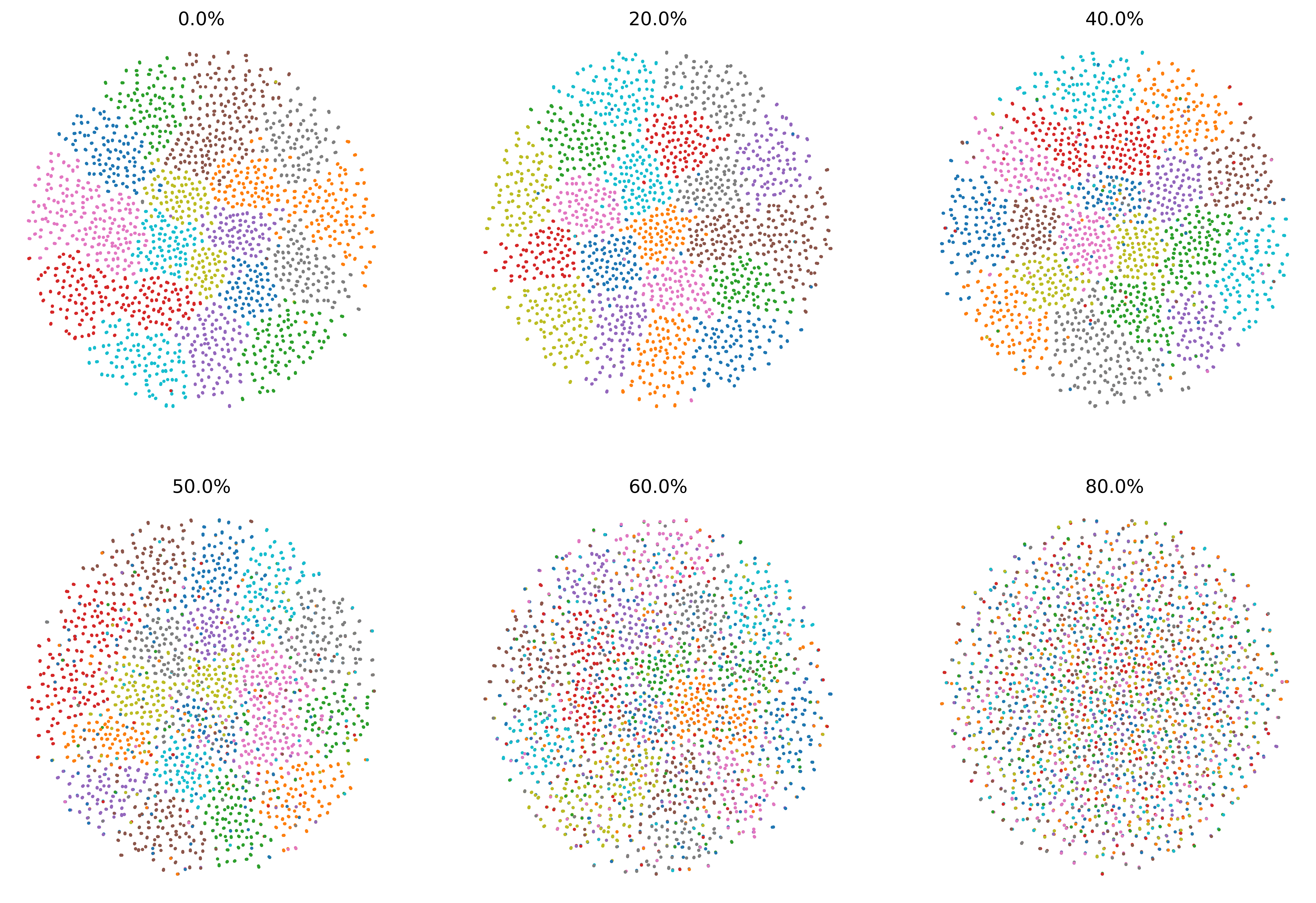}
 \caption{Different degrees of mode alignments of datasets $X$ visualized using TSNE.
 The alignment degree corresponds to the angle between modes. $0 \%$ alignment corresponds to orthogonal modes and $80 \%$ corresponds to an angle of $18^\circ$.}
 \label{fig:tsne}
\end{figure}

Figure \ref{fig:tsne} visualizes datasets $X$ with different degrees of mode alignment.

\subsection{InfoNCE with soft targets: discussion} \label{supp:snce_discussion}

To simplify notation, we set $\alpha_k = [p_{Y|X}(1|x), \dots, p_{Y|X}(K|x)]$ and $\alpha_l=[\eta_{l1}, \dots, \eta_{lK}]$, $l\neq k$.
Attraction-repulsion in 
\begin{align} \label{eq:energyXent_}
    l = - \sum_{i=1}^K \alpha_{ki} \log
      \frac{
    \exp \left( s(z, y_i; \tau, \eta) \right)
    }{
    \sum_{l=1}^{N+1} \exp \left( \sum_{j=1}^K \alpha_{lj} s(z, y_j; \tau, \eta) \right)
    },
\end{align}
can be further separated by restating the loss function as the two summands
\begin{align} \label{eq:attr_energyXent}
    - \sum_{i=1}^K \alpha_{ki} \log
      \frac{
    \exp \left( s(z, y_i) \right)
    }{
    \exp(\alpha_{ki} s(z, y_k)) + \sum_{\substack{l=1 \\ l \neq k}}^{N+1} \exp \left( \alpha_{li} s(z, y_i) + \sum_{\substack{j=1 \\ j \neq i}}^K \tilde{\alpha}_{lj} s(z, y_j) \right)
    },
\end{align}
with $\tilde{\alpha}_{lj} = \alpha_{lj} - \alpha_{ki}$, and
\begin{align} \label{eq:rep_energyXent}
    \sum_{i=1}^K \alpha_{ki} \sum_{\substack{l=1 \\ l \neq k}}^K \alpha_{li} s(z, y_l),
\end{align}
where we have omitted the dependence of the scoring function $s$ on $\tau$ and $\eta$ for readability.
Similar to \eqref{eq:energyXent_}, \eqref{eq:attr_energyXent} predominantly contributes with an attractive pull between the label embedding $y_i$ and the data embedding $z$ that is regulated by $\alpha_{ki}$, due to the numerator in the logarithm.
We discuss the denominator as well as the second term \eqref{eq:rep_energyXent} in the context of label smoothing with a uniform noise distribution $\xi \equiv 1 / K$, i.e. $\alpha_{k i_0} = 1 - (K-1)\epsilon / K$ for the true class $i_0$ and $\alpha_{kj} = \epsilon / K$ for $j \neq i_0$ otherwise.
For the composition of noise labels, we differentiate between the following cases:
\begin{itemize}
    \item In case of $i=i_0$ and the $l$-th sample has label $i_l \neq i_0$, we have $\alpha_{li} = \epsilon / K, \tilde{\alpha}_{li_l} = 1 - \epsilon$, and $\tilde{\alpha}_{lj}=0$ otherwise. Effectively this acts as repulsion between $z$ and label $i_0$ of strength $\epsilon / K$ as well as $z$ and label $i_l$ of strength $1 - \epsilon$.

    \item In case of $i = i_0$ and the $l$-th sample has label $i_l = i_0$, we have $\alpha_{li_0} = 1 - (K-1)\epsilon / K$, and $\tilde{\alpha}_{lj} = 0$, resulting in a repulsion between $z$ and $i_0$.
    
    \item In case of $i \neq i_0$ and the $l$-th sample has label $i_l \neq i_0$, we have $\alpha_{li_0} = \epsilon - 1$ which results in an attractive force between $z$ and label $i_0$.
    Further, in case of $i_l = i$ then $\alpha_{li} = 1 - (K-1) \epsilon / K$, and $\tilde{\alpha}_{lj} = 0$ otherwise.
    In the event of $i_l \neq i$, we have $\alpha_{li}= \epsilon / K, \tilde{\alpha}_{li_l} = 1 - \epsilon$, and $\tilde{\alpha}_{lj} = 0$ otherwise.
    Ultimately, this acts as a repulsion between $z$ and the labels $i, i_l$ with the respective strength.
    

    \item In case of $i \neq i_0$ and the $l$-th sample has label $i_l = i_0$, we have $\alpha_{li} = \epsilon / K$ and $\tilde{\alpha}_{lj} = 0$ otherwise, which results in a repulsive force between $z$ and label $i$.
\end{itemize}
The second term~\eqref{eq:rep_energyXent} is a penalty term and acts as a repulsion between $z$ and all classes while penalzing the true class $i_0$ less than others in case of $\epsilon < 1/2$.
In summary, soft targets change the attraction repulsion dynamics between an embedded data point and all label embeddings by adjusting the amount of pull-push applied.

\paragraph{Gradient analysis.}
Here, we derive and examine the gradient responses of soft target InfoNCE with respect to data embeddings $z$.
We consider as scoring function the cosine similarity (with temperature $\tau=1$) $s(z,y)= z^Ty / \|z\|\|y\|$ and first compute the gradient with respect to $\hat{z} = z / \|z\|$.
\begin{align}
    \partial_{\hat{z}} l &= - 
    \left(
    \sum_{i=1}^K \alpha_{ki} y_i -
    \sum_{l=1}^{N+1} C_l(z,y)
    \sum_{j=1}^{K} \alpha_{lj} y_j
    \right)\\
    &= \sum_{i=1}^K
    \left(
    - \alpha_{ki} y_i +
    \sum_{l=1}^{N+1} C_l(z,y) \alpha_{lj}
    \right) y_j,
\end{align}
with $C_l(z,y) = \exp\left(\sum_{j=1}^K \alpha_{lj} s(z, y_j\right) / \sum_{l=1}^{N+1} \exp\left(\sum_{j=1}^K \alpha_{lj} s(z, y_j)\right)$.
Thus, for the unnormalized embeddings $z$, we have
\begin{align}
    \partial_z l = \frac{I - \hat{z} \hat{z}^T}{\|z\|} \partial_{\hat{z}} l
\end{align}

For soft target embeddings, our loss function inherits gradient response properties directly from the original InfoNCE.
Thus, for weak targets\footnote{These are targets with $z^T \cdot y \approx 1$ in the case of true targets, and $z^T \cdot y \approx -1$ in the case of noise or negative targets.} gradient magnitudes approach zero, while for strong targets\footnote{These are targets with $z^T \cdot y \approx 0$.} gradient magnitudes are larger than zero.

For the original hard targets, we obtain gradient magnitudes larger than zero when the target is strong, \emph{i.e.} when $z^T \cdot y_j \approx 0$.
More specifically, we would obtain the following gradient response with respect to $y_j$:
\begin{align} \label{eq:gradient_magnitude}
    \alpha_{kj} \left\| -1 + \frac{\sum_{l=1}^{N+1} \alpha_{lj} \exp\left( \sum_{i=1}^K \alpha_{li} s(z, y_i)\right)}{\sum_{i=1}^K \alpha_{kj} \exp\left(\sum_{i=1}^K \alpha_{li} s(z,y_i)\right)} \right\|.
\end{align}
As in the previous discussion, we consider the case of label smoothing with a uniform noise distribution.
In this case, for $\epsilon \in (0, 1 / 2)$ the second sum in the denominator is smaller than $1$ since $\alpha_{lj} / \alpha_{kj} \leq 1$, and larger otherwise.
Thus, in both cases the term in~\eqref{eq:gradient_magnitude} is larger than zero.

The analytic expression of soft target InfoNCE gradients is similar to those of soft target cross-entropy, although with distinct characteristics:
soft target InfoNCE involves a summation over batch sample embeddings instead of classes, and additionally, the soft labels directly impact the logits.
The analytical expressions we derived for the gradient magnitudes for scenarios involving hard positive and hard negative pairs reveal that in the presence of hard positives, the gradient magnitude increases with the number and strength of the negative pairs.
Similarly, in instances in which hard negatives are given, the gradient magnitude increases proportionally with the strength of the positive as well as the number and strength of the negative pairs.

\subsection{InfoNCE with soft distributions} \label{supp:sdinfonce}

\paragraph{Derivation.}
Let $S$ be the random variable denoting the position of $y_0$ within $T$.
We are interested in recovering the position $S$ given the knowledge of $T$, \emph{i.e.} $P(S|T)$.
A priori, we have $P(S=i) = 1 / (N+1)$ for all allowed positions $i = 0, \dots, N$.
Further, we know that $P(y_j | S=0) = \eta(y_j)$ for all $j \neq 0$, and $P(y_0 | S=0) = p^\epsilon(y_0)$. Thus,
\begin{align} \label{eq:sdinfonce_deriv1}
    P(T|S=0) = \frac{p^\epsilon(y_0|x)}{\eta(y_0)} \prod_{i=0}^N \eta(y_i).
\end{align}
From this, we can infer
\begin{align} \label{eq:sdinfonce_deriv2}
    P(T) = \frac{1}{N+1} \prod_{i=0}^N \eta(y_i) \sum_{k=0}^N \frac{p^\epsilon(y_k|x)}{\eta(y_k)}.
\end{align}
By applying Bayes' theorem in combination with~\eqref{eq:sdinfonce_deriv1} and~\eqref{eq:sdinfonce_deriv2}, we can compute the posterior
\begin{align} \label{eq:sdinfonce_posterior}
    P(S=0 | T) = \frac{\frac{p^\epsilon(y_0|x)}{\eta(y_0)}}{\sum_{k=0}^N\frac{p^\epsilon(y_k|x)}{\eta(y_k)}}.
\end{align}
We estimate the ratio $p^\epsilon(y|x) / \eta(y)$ using the parameterized model
\begin{align} \label{eq:parametric_model}
    \exp(s(z, y) / \tau),
\end{align}
where $z=f(x; \theta)$ is for instance a neural network.
Inserting~\eqref{eq:parametric_model} into~\eqref{eq:sdinfonce_posterior} and taking the negative logarithm results in the soft distribution InfoNCE loss function
\begin{align}
    l^{\mathrm{SDInfoNCE}}(T) = 
    -\log \frac{\exp(s(z, y_0) / \tau)}{\sum_{k=0}^N \exp(s(z, y_k) / \tau)}.
\end{align}
Notably, for $\eta \equiv \xi$, the parameterized model~\eqref{eq:parametric_model} converges to a function proportional to the density ratio
\begin{align}
    (1 - \epsilon) \frac{p(y|x)}{\xi(y)} + \epsilon.
\end{align}
Multiplying by $\xi(y)$ allows to recover the density $(1 - \epsilon) p(y|x) + \epsilon \xi(y)$.

\paragraph{Effect of soft distributions on the MI lower bound: derivation.}

Here we derive a variational lower bound for soft distribution InfoNCE connecting MI and entropy properties of the learned representations.
To recapitulate, let $p_{Y, Z}$ be the "data" distribution, $p_Y$ the negative sampling distribution as well as the noise distribution for label smoothing, \emph{i.e.} $p_Y \equiv \eta \equiv \xi$.
Further, we define the conditional and joint soft data distributions as
\begin{align}
    p_{Y|Z}^\epsilon(y|z) &= (1-\epsilon) p_{Y|Z}(y|z) + \epsilon p_Y(y), \\
    p_{Z,Y}^\epsilon(z, y) &= (1-\epsilon) p_{Z,Y}(z, y) + \epsilon p_Y(y)p_Z(z).
\end{align}

Next, we derive the lower bound
\begin{align}
    l^{\mathrm{SDInfoNCE}}(T) &\geq \underset{T}{\mathbb{E}} \left[- \log \frac{\frac{p_{Y|Z}^\epsilon(y_0|x)}{p_Y(y_0)}}{\sum_{k=0}^N\frac{p_{Y|Z}^\epsilon(y_k|x)}{p_Y(y_k)}}\right] \\
    & = \underset{T}{\mathbb{E}} \left[ \log \left(
    1 + \frac{p_Y(y_0)}{p_{Y|Z}^\epsilon(y_0|x)} \sum_{k=1}^N \frac{p_{Y|Z}^\epsilon(y_k|x)}{p_Y(y_k)} \right)
    \right] \\
    & \approx
    \underset{T}{\mathbb{E}} \left[ \log \left(
    1 + \frac{p_Y(y_0)}{p_{Y|Z}^\epsilon(y_0|x)} N \underset{y \sim p_Y}{\mathbb{E}}
    \left[\frac{p_{Y|Z}^\epsilon(y|x)}{p_Y(y)}\right] \right)
    \right] \\
    &= \underset{T}{\mathbb{E}} \left[ \log \left(
    1 + \frac{p_Y(y_0)}{p_{Y|Z}^\epsilon(y_0|x)} N \right)
    \right] \\
    & \geq
    \log N -  \underset{(z,y) \sim p_{Z,Y}^\epsilon}{\mathbb{E}} \left[ \log \frac{p_{Y|Z}^\epsilon(y|z)}{p_Y(y)} \right], \label{supp:prebound}
\end{align}
where $N$ denotes the the number of negative samples.

We can rewrite the expectation in \eqref{supp:prebound} into
\begin{align}
    &\underset{(z,y) \sim p_{Z,Y}^\epsilon}{\mathbb{E}} \left[ \log \frac{(1-\epsilon)p_{Y|Z}(y|z) + \epsilon p_Y(y)}{p_Y(y)} \right] \\
    &=
     \underset{(z,y) \sim p_{Z,Y}^\epsilon}{\mathbb{E}} \left[ \log \frac{p_{Y|Z}(y|z)}{p_Y(y)} \right]
    +  \underset{(z,y) \sim p_{Z,Y}^\epsilon}{\mathbb{E}} \left[ \log \left(1 - \epsilon + \epsilon \frac{p_Y(y)}{p_{Y|Z}(y|z)} \right) \right] \\
    &=
    (1-\epsilon) \underset{(z, y) \sim p_{Z,Y}}{\mathbb{E}} 
    \left[
    \log \frac{p_{Y|Z}(y|z)}{p_Y(y)}
    \right]
    + \epsilon \underset{\substack{z \sim p_Z \\ y \sim p_Y}}{\mathbb{E}}
    \left[
    \log \frac{p_{Y|Z}(y|z)}{p_Y(y)}
    \right]\\
    &+ \underset{(z,y) \sim p_{Z,Y}^\epsilon}{\mathbb{E}} \left[ \log \left(
    \frac{(1 - \epsilon) p_{Y|Z}(y|z) + \epsilon p_Y(y)}{p_{Y|Z}(y|z)} \right) \right]
    \\
    &=
    (1-\epsilon) \left( I(Z;Y)
    + \underset{(z,y) \sim p_{Z,Y}}{\mathbb{E}} \left[ \log \left(
    \frac{(1 - \epsilon) p_{Y|Z}(y|z) + \epsilon p_Y(y)}{p_{Y|Z}(y|z)} \right) \right] \right) \\
    & + \epsilon \left( \underset{\substack{z \sim p_Z \\ y \sim p_Y}}{\mathbb{E}}
    \left[ \log \frac{p_{Y|Z}(y|z)}{p_Y(y)} \right]
    + \underset{\substack{z \sim p_Z \\ y \sim p_Y}}{\mathbb{E}}
    \left[ \log \left(
    \frac{(1 - \epsilon) p_{Y|Z}(y|z) + \epsilon p_Y(y)}{p_{Y|Z}(y|z)} \right) \right] \right) \\
    &=
    (1-\epsilon)\left( I(Z;Y)
    + \underset{(z,y) \sim p_{Z,Y}}{\mathbb{E}} \left[ \log \frac{p_{Y|Z}^\epsilon(y|z)}{p_{Y|Z}(y|z)} \right]
    \right)
    + \epsilon \underset{\substack{z \sim p_Z \\ y \sim p_Y}}{\mathbb{E}}
    \left[ \log \frac{p_{Y|Z}^\epsilon(y|z)}{p_Y(y)} \right] \\
    &=
    (1-\epsilon)\left( I(Z;Y) + H(Y|Z)
    \right)
    + \underset{(z,y) \sim p_{Z,Y}^\epsilon}{\mathbb{E}} \left[ \log p_{Y|Z}^\epsilon(y|z) \right]
    + \epsilon H(Y) \\
    &=
    (1-\epsilon)\left( I(Z;Y) + H(Y|Z) \right)
    + \underset{(z,y) \sim p_{Z,Y}^\epsilon}{\mathbb{E}} \left[ \log p_{Y|Z}^\epsilon(y|z) \right] \\
    &+ \epsilon \left( I(Z;Y) + H(Y|Z) \right)
    \\
    &= \label{supp:mi_deriv_1}
    I(Z;Y) + H(Y|Z)
    + \underset{(z,y) \sim p_{Z,Y}^\epsilon}{\mathbb{E}} \left[ \log p_{Y|Z}^\epsilon(y|z) \right].
\end{align}
To simplify notation we define
\begin{align}
    H_\epsilon(Y|Z):=
    \underset{(z,y) \sim p_{Z,Y}^\epsilon}{\mathbb{E}} \left[ - \log p_{Y|Z}^\epsilon(y|z) \right].
\end{align}
Inserting \eqref{supp:mi_deriv_1} back into \eqref{supp:prebound}, we obtain
\begin{align} \label{supp:softMIbound}
    I(Z;Y) + H(Y|Z) - H_\epsilon(Y|Z) \geq \log N - l^{\mathrm{SDInfoNCE}}(T).
\end{align}
Next we show that the l.h.s. of \eqref{supp:softMIbound} is larger or equal to zero:
\begin{align}
    H_\epsilon(Y|Z) &= H_\epsilon(Y,Z) - H(X) \\
    &\leq H(Y) + H(X) - H(X) = H(Y).
\end{align}
Since $H(Y) = I(Z;Y) + H(Y|Z)$, we have $I(Z;Y) + H(Y|Z) - H_\epsilon(Y|Z) \geq 0$.

\subsection{Experiments} \label{supp:experiments}

\subsubsection{Classification accuracy}
Table \ref{tab:top_5} shows top-5 classification accuracy for ImageNet, Tiny ImageNet, CIFAR-100 and class-average accuracy for CellTypeGraph for all baselines.

Here we provide further technical details for our classification experiments.

\paragraph{Image classification.} Table \ref{tab:vit_params} details the settings used across all baselines.
As classification head we used a $3$-layer MLP with hidden dimension $4096$.
On ImageNet, we used as base learning rate of $0.00055$ for InfoNCE-type loss functions, and $0.0003$ for cross-entropy-type baselines.
On Tiny ImageNet, we used $0.00025$ for InfoNCE-type baselines, and $0.0003$ for cross-entropy-type baselines.
On CIFAR-100, we used $0.0003$ for InfoNCE, soft distribution InfoNCE and cross-entropy-type baselines, and $0.00055$ for soft target InfoNCE.
Note, we determined (base) learning rates based on a search over multiple samples from the interval $[0.00005, 0.003]$.

\paragraph{Node classification.}
Besides the parameter settings we described in the main experiments, all DeeperGCN models were trained with adjacency dropout (dropout prob. of $0.1$), adding of edges (ratio of 0.1), node feature dropout and masking (prob. of $0.1$), random normal noise ($\sigma$ of $0.1$) and random basis transforms.

\begin{table}[tb]
  \caption{Top-5/class average classification accuracy for various datasets. Comparison of four baseline loss functions to soft target InfoNCE.
  }
  \label{tab:top_5}
  \centering
  \begin{tabular}{@{}lcccc@{}}
    \toprule
     & ImageNet \ \ & Tiny ImageNet \ \ & CIFAR-100 \ \ & CellTypeGraph\\
    \midrule
    NLL & 94.95 & 93.84 & 98.37 & 80.57 \\
    Soft target cross-entropy & \textbf{96.46} & 94.91 & 98.40 & \textbf{81.41} \\
    InfoNCE & 95.43 & 93.67 & 98.40 & 80.00 \\
    \midrule
    Soft distribution InfoNCE & 96.40 & \textbf{95.06} & \textbf{98.78} & 79.97 \\
    Soft target InfoNCE  & 96.43 & 94.91 & 98.51  & 81.27 \\
  \midrule
  & top-5 acc. & top-5 acc. & top-5 acc. & class-avg. acc.\\
  \bottomrule
  \end{tabular}
\end{table}

\begin{table}[tb]
  \caption{Parameter settings used to train ViT-B/16 models across all baselines.
  }
  \label{tab:vit_params}
  \centering
  \begin{tabular}{@{}ll@{}}
    \toprule
     parameter \hspace{2.5cm} & value \\
    \midrule
     optimizer & AdamW \\
     weight decay & 0.05 \\
     layer-wise lr decay & 0.65 \\
     batch size & 1024 \\
     learning rate schedule & cosine decay \\
     warmup epochs & 5 \\
    training epochs & 100 \\
    label smoothing & 0.1 \\
    MixUp & 0.8 \\
    CutMix & 1.0 \\
    augmentations & RandAug(9, 0.5)\\
  \bottomrule
  \end{tabular}
\end{table}

\subsubsection{Ablation experiments}

\begin{figure}[t]
 \centering
 \includegraphics[width=8.5cm]{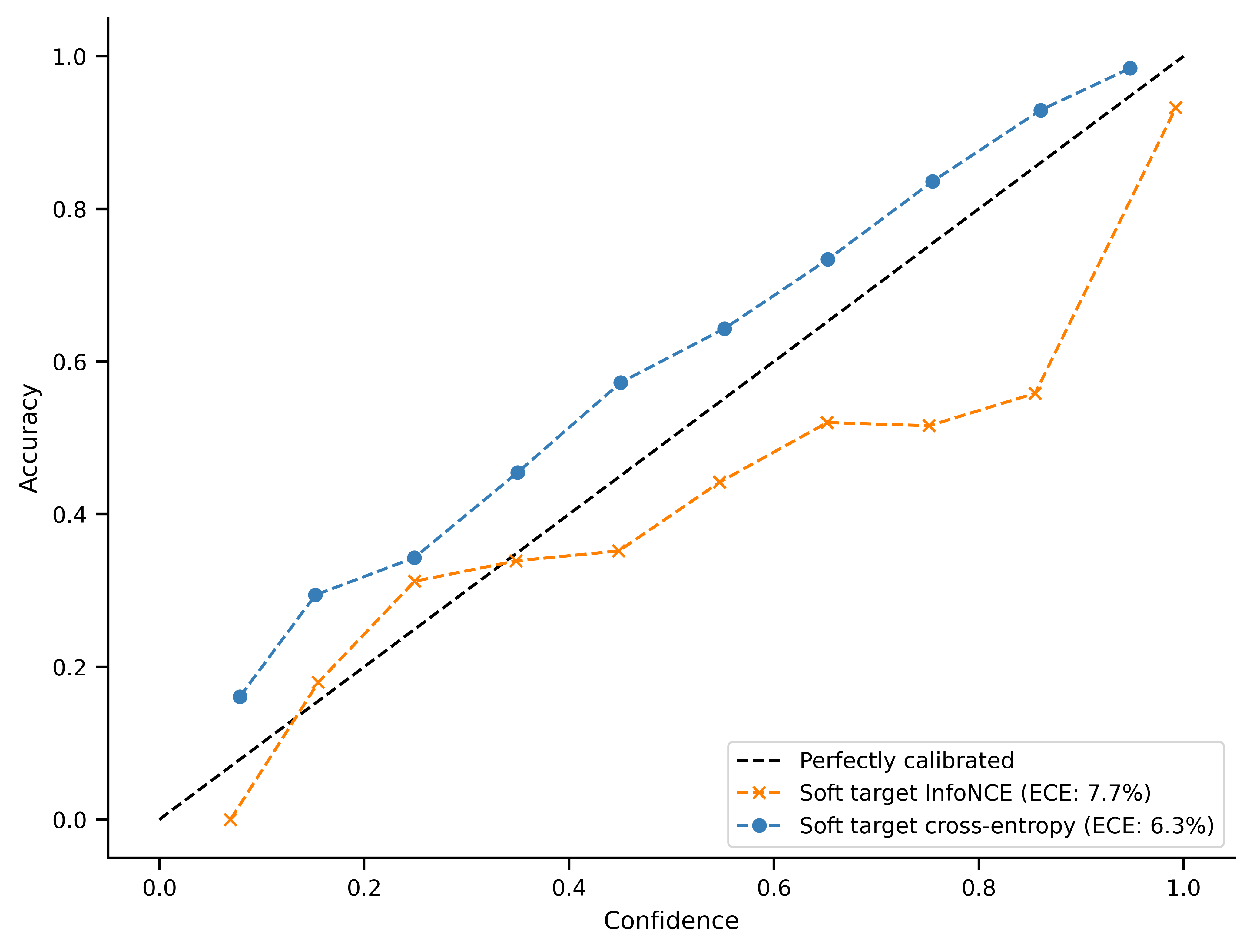}
 \caption{Reliability diagram of the best performing ViT-B/16 models on Tiny ImageNet, trained with soft target loss functions in combination with label smoothing, MixUp and CutMix.}
 \label{fig:reliability_best_model_tiny_imagenet}
\end{figure}

\paragraph{Confidence calibration.} Figure 
\ref{fig:reliability_best_model_tiny_imagenet} shows the reliability diagram of the best performing ViT-B/16 models on Tiny ImageNet.

\end{document}